\documentclass[11pt, a4paper, onecolumn, copyright]{AweAI}

\usepackage{multirow}
\usepackage[authoryear, round]{natbib} % sort&compress
\usepackage{graphicx}
\usepackage{microtype}
\usepackage{subcaption}
\usepackage{booktabs}

\usepackage{listings} 
\usepackage{makecell}

\usepackage[most]{tcolorbox}
\tcbuselibrary{listings}
\usepackage{xspace}
\usepackage{textcomp}
\usepackage{algorithm}
\usepackage{algpseudocode}
\usepackage{amsmath}
\usepackage{amssymb}
\usepackage{mathtools}
\usepackage{amsthm}
\usepackage{enumitem}
\usepackage{pifont}
\definecolor{darkgreen}{RGB}{50,100,0}
\definecolor{darkred}{RGB}{200, 0, 0}

\usepackage{xurl}
\usepackage{titletoc}
\usepackage{needspace}

\usepackage{float}       % figure/table [H]
\usepackage{placeins}    % \FloatBarrier
\usepackage{tabularx}    % tabularx
\usepackage{array}       % custom table columns and \arraybackslash
\usepackage{marvosym}

\definecolor{tableblue}{HTML}{E6EFFA}
\definecolor{calibrationorange}{HTML}{FFF0E3}
\definecolor{headergray}{HTML}{444444}

\colorlet{DarkGreen}{green!50!black}
\colorlet{DarkRed}{red}

\DeclareRobustCommand{\github}{%
  \raisebox{-1.5pt}{%
    \includegraphics[height=1.05em]{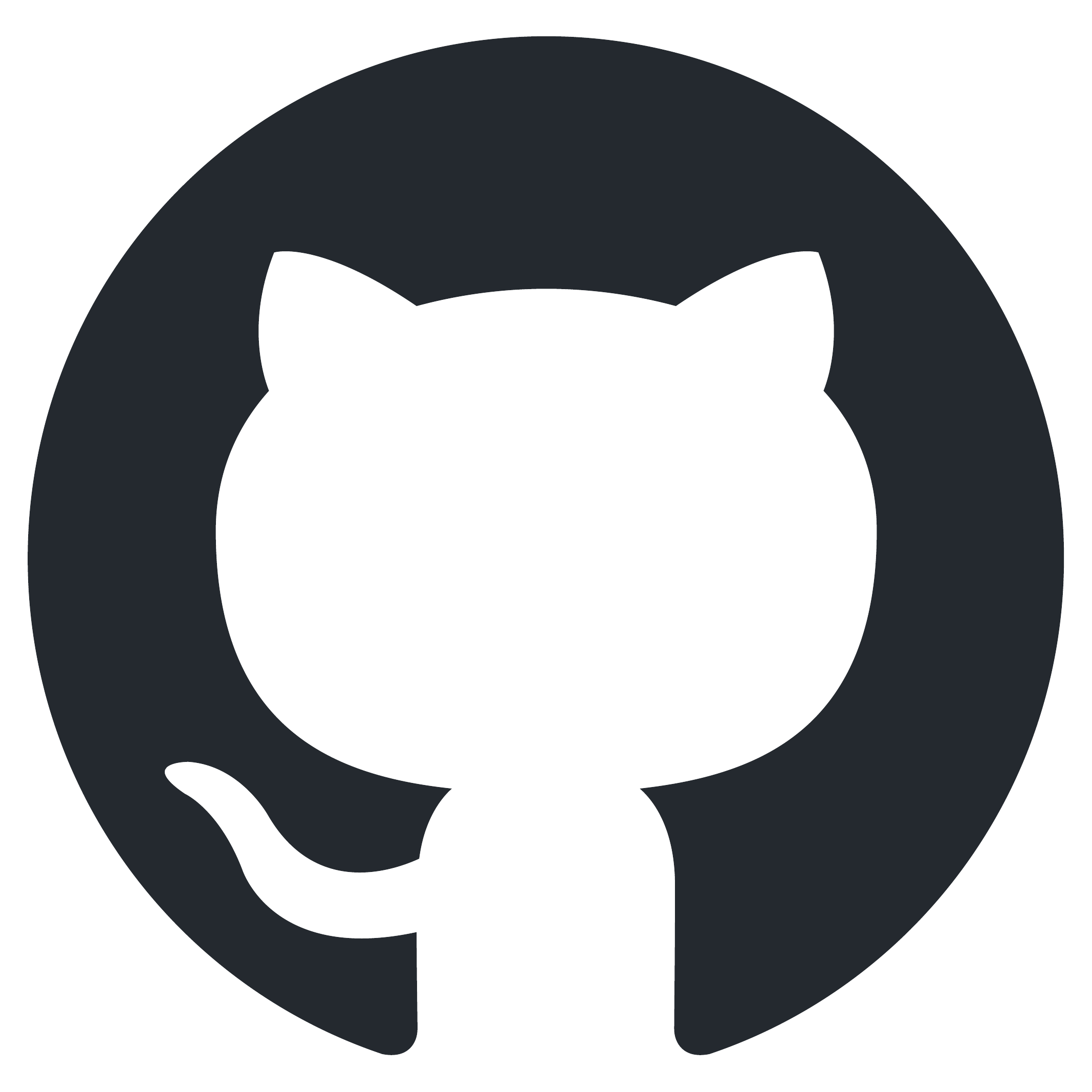}%
  }\xspace
}

\DeclareRobustCommand{\huggingface}{%
  \raisebox{-1.5pt}{%
    \includegraphics[height=1.05em]{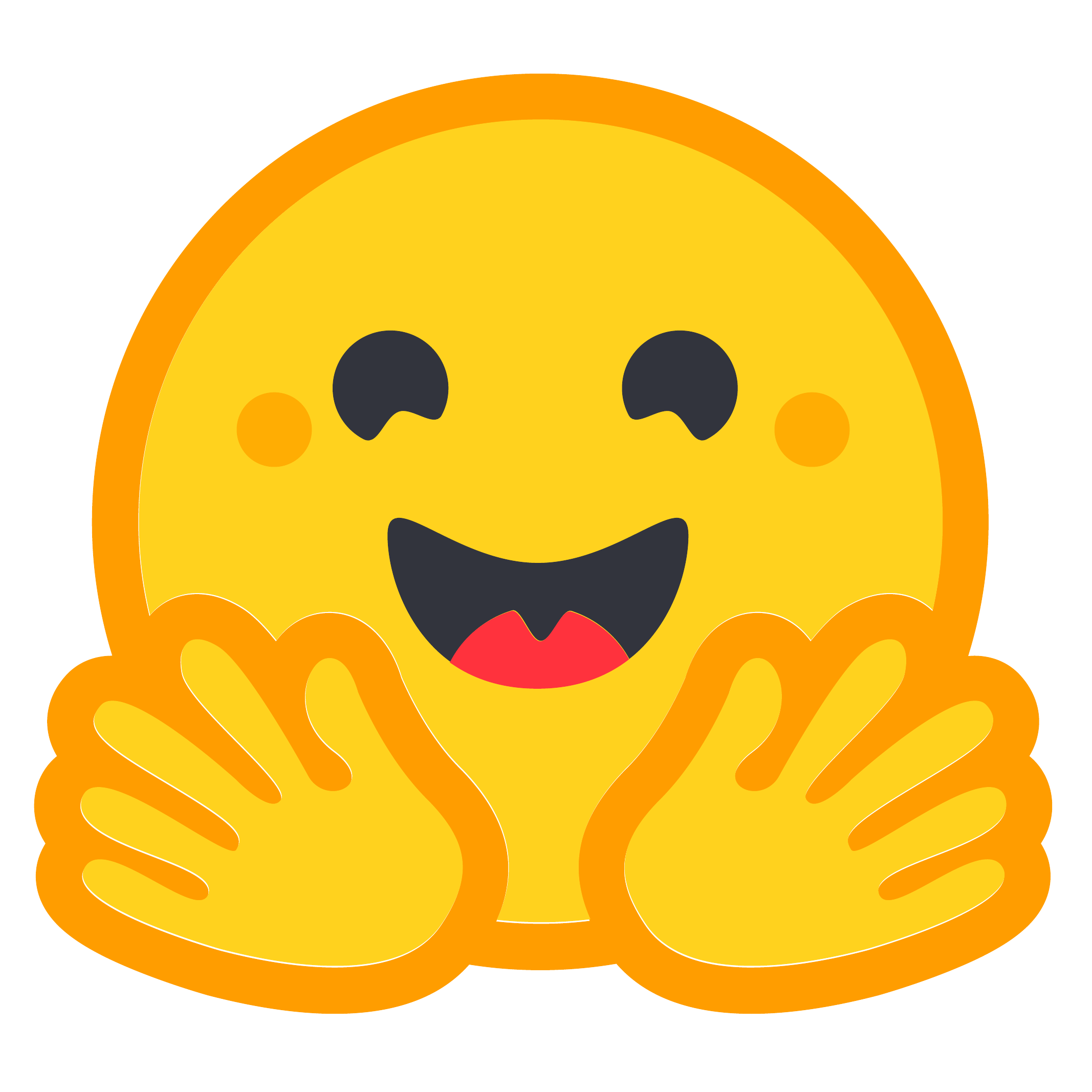}%
  }\xspace
}

\DeclareRobustCommand{\page}{%
  \raisebox{-1.5pt}{%
    \includegraphics[height=1.05em]{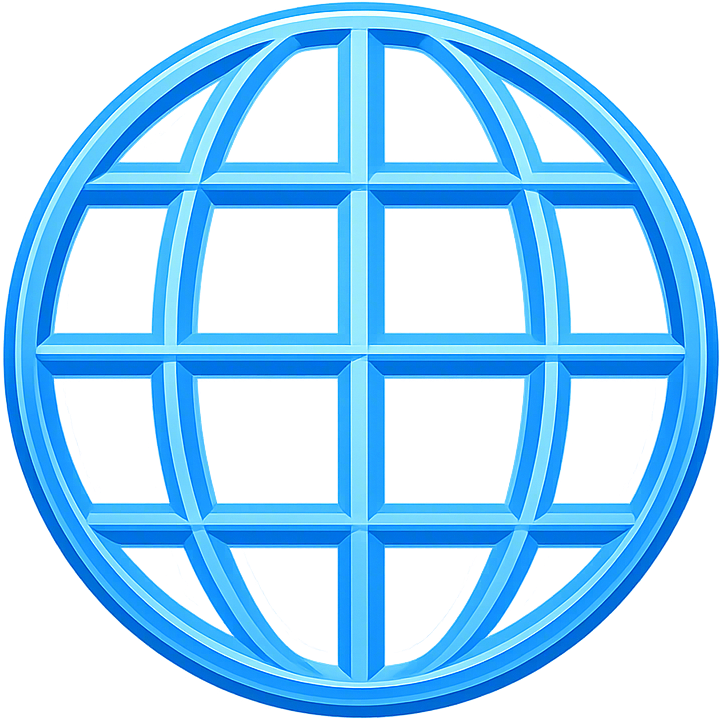}%
  }\xspace
}

\newcommand{\resourcebuttons}{%
  \page\
  \href{https://aweai-team.github.io/AutoMaster/}{Project Page}%
  \hspace{1.5em}%
  \huggingface\
  \href{https://huggingface.co/collections/RUC-AIBOX/automaster}{Models}%
  \hspace{1.5em}%
  \github\
  \href{https://github.com/RUCAIBox/AutoMaster}{GitHub}%
}

\definecolor{caseblue}{HTML}{3F6288}
\definecolor{caseblueback}{HTML}{F2F7FF}
\definecolor{casegreen}{HTML}{397A57}
\definecolor{casegreenback}{HTML}{F2F8F4}
\definecolor{casered}{HTML}{8F3F3F}
\definecolor{casegray}{HTML}{6F7477}
\definecolor{casecontrast}{HTML}{C46A6A}
\definecolor{casemulti}{HTML}{5D7FA6}
\definecolor{caseorange}{HTML}{D7A23A}
\definecolor{caseorangeback}{HTML}{FFF8E8}
\definecolor{casegrayback}{HTML}{F4F5F5}

\newtcolorbox{takeawaybox}{
  enhanced,
  colback=caseblueback,
  colframe=caseblueback,
  boxrule=0pt,
  arc=1.5mm,
  left=2.5mm,
  right=2.5mm,
  top=1.5mm,
  bottom=1.5mm,
  before skip=5pt,
  after skip=7pt,
  fontupper=\small\slshape
}

\lstdefinestyle{promptstyle}{
  basicstyle=\footnotesize\ttfamily,
  breaklines=true,
  breakatwhitespace=false,
  columns=fullflexible,
  keepspaces=true,
  showstringspaces=false,
  tabsize=2,
  literate={–}{{--}}1 {—}{{---}}1 {…}{{...}}3
           {→}{{$\to$}}1 {↔}{{$\leftrightarrow$}}1
           {─}{{-}}1 {│}{{|}}1 {└}{{+}}1 {├}{{+}}1
           {✓}{{[PASS]}}6 {✗}{{[FAIL]}}6
}

\newtcblisting{promptbox}[2][]{
  enhanced,
  breakable,
  listing only,
  listing engine=listings,
  listing options={style=promptstyle},
  colback=gray!3,
  colframe=gray!70!black,
  colbacktitle=gray!70!black,
  coltitle=white,
  title={#2},
  fonttitle=\bfseries,
  boxrule=0.7pt,
  arc=1.5mm,
  left=4mm,
  right=4mm,
  top=2mm,
  bottom=2mm,
  before skip=8pt,
  after skip=10pt,
  #1
}

\newtcolorbox{caseinput}[1]{
  enhanced, breakable, colback=caseblueback, colframe=caseblue,
  colbacktitle=caseblue, coltitle=white, title={#1}, fonttitle=\bfseries,
  boxrule=0.7pt, arc=1mm, left=4mm, right=4mm, top=2.5mm,
  bottom=2.5mm, before skip=5pt, after skip=4pt
}

\newtcolorbox{casesearch}[1]{
  enhanced, breakable, colback=white, colframe=caseblue,
  colbacktitle=caseblue, coltitle=white, title={#1}, fonttitle=\bfseries,
  boxrule=0.7pt, arc=1mm, left=3mm, right=3mm, top=2.5mm,
  bottom=2.5mm, before skip=5pt, after skip=4pt
}

\newtcolorbox{caseoutput}[1]{
  enhanced, breakable, colback=white, colframe=caseblue,
  colbacktitle=caseblue, coltitle=white, title={#1}, fonttitle=\bfseries,
  boxrule=0.7pt, arc=1mm, left=4mm, right=4mm, top=2.5mm,
  bottom=2.5mm, before skip=5pt, after skip=4pt
}

\newtcolorbox{casecalibration}[1]{
  enhanced, breakable, colback=casegreenback, colframe=casegreen,
  colbacktitle=casegreen, coltitle=white, title={#1}, fonttitle=\bfseries,
  boxrule=0.7pt, arc=1mm, left=4mm, right=4mm, top=2.5mm,
  bottom=2.5mm, before skip=5pt, after skip=4pt
}

\newtcolorbox{revisioncasebox}[2]{
  enhanced, breakable, colback=white, colframe=#2,
  colbacktitle=#2, coltitle=white, title={#1}, fonttitle=\bfseries,
  boxrule=0.8pt, arc=1mm, left=4mm, right=4mm, top=2.5mm,
  bottom=2.5mm, before skip=6pt, after skip=5pt
}

\newtcblisting{markdownbox}[1][]{
  listing engine=listings,
  listing options={
    basicstyle=\scriptsize\ttfamily, % 保持与你原来一致的 \scriptsize
    breaklines=true,              % 自动换行
    breakatwhitespace=true,
    numbers=none,                 % 不显示行号
    columns=fullflexible,         % 优化对齐
    keepspaces=true,              % 保留代码中的空格
    aboveskip=0pt,
    belowskip=0pt
  },
  colback=gray!5!white,           % 背景色：极浅灰
  colframe=gray!70!black,         % 边框色：深灰
  title={\textbf{Prompt (Markdown)}}, % 默认标题
  fonttitle=\bfseries,
  listing only,                   % 只显示代码，不运行
  left=5mm, right=5mm, top=3mm, bottom=3mm, % 内边距
  breakable,                      % 允许跨页
  enhanced,                       % 启用高级外观
  overlay={\begin{tcbclipinterior}\fill[gray!5!white] (frame.south west) rectangle ([xshift=0mm]frame.north west);\end{tcbclipinterior}}, % 纯净背景
  #1                              % 允许修改参数
}

\newtcbinputlisting{\markdownfile}[2][]{
  listing engine=listings,
  listing file={#2}, % 读取的文件路径
  listing options={
    basicstyle=\scriptsize\ttfamily,
    breaklines=true,
    breakatwhitespace=true,
    numbers=none,
    columns=fullflexible,
    keepspaces=true
  },
  colback=gray!5!white,
  colframe=gray!70!black,
  title={\textbf{Prompt (Markdown)}},
  fonttitle=\bfseries,
  listing only,
  breakable,
  enhanced,
  #1
}

\usepackage{xcolor}
\usepackage{fancyvrb}
\usepackage{tcolorbox}
\tcbuselibrary{skins,breakable}

\definecolor{AMSkillInk}{HTML}{244F65}
\definecolor{AMSkillLine}{HTML}{B8CAD5}
\definecolor{AMSkillTint}{HTML}{EDF3F7}
\definecolor{AMSkillPaper}{HTML}{FAFCFD}
\definecolor{AMSkillMuted}{HTML}{576674}

\newtcolorbox[auto counter,number within=section]{AMskillbox}[3][]{
  enhanced,
  breakable,
  colback=AMSkillPaper,
  colframe=AMSkillLine,
  colbacktitle=AMSkillTint,
  coltitle=AMSkillInk,
  title={%
    \strut Skill~\thetcbcounter\quad
    {\mdseries\ttfamily #2}\hfill
    {\normalfont\rmfamily\footnotesize #3}%
  },
  fonttitle=\normalfont\rmfamily\bfseries\small,
  fontupper=\normalfont\rmfamily\small,
  boxrule=0.5pt,
  arc=1.3mm,
  left=9pt,
  right=9pt,
  top=7pt,
  bottom=7pt,
  boxsep=0pt,
  before skip=6pt,
  after skip=13pt,
  before upper={%
    \setlength{\parindent}{0pt}%
    \setlength{\parskip}{3.5pt}%
    \fvset{%
      listparameters={%
        \setlength{\topsep}{3pt}%
        \setlength{\partopsep}{0pt}%
      }%
    }%
    \linespread{1.025}\selectfont
  },
  #1
}

\newcommand{\AMskillOmission}{%
  \noindent{\color{AMSkillMuted}\texttt{[...]}}\enspace
}

\uselogo{}

\newcommand{\papername}{AutoMaster}
\newcommand{\method}{Skill-FT}

\title{\papername{}: Autonomous Post-Training by Making the Same Model a Better Teacher}

\newcommand{\publicday}{Nov.~20, 2026}

\author[*]{Guoxin Chen}
\author[*]{Shuang Sun}
\author[*]{Fanzhe Meng}
\author[ \hspace{-0.3em}]{Xinyu Tang}
\author[ \hspace{-0.3em}\Letter]{\authorcr Ruihua Song}
\author[ \hspace{-0.3em}\Letter]{Wayne Xin Zhao}
\footerlinks{\resourcebuttons}

\correspondingauthor{\{gx.chen.chn, sunshuanguns, mengfanzhe16, batmanfly\}@gmail.com}

\affil[1]{Gaoling School of Artificial Intelligence, Renmin University of China\footnote{$^*$Equal Contributions. \textsuperscript{\Letter}Corresponding authors. \hfill \textbf{Date:} \publicday.}}
\begin{abstract}
Task success alone does not specify which behaviors post-training
should reinforce.
Even successful teacher trajectories can contain unsupported
assumptions, unproductive repetition, or superficial verification.
We propose \method{} (Skill-Guided Fine-Tuning), which uses
natural-language skills to suppress undesirable habits and elicit
underrepresented effective behaviors during teacher rollouts,
then fine-tunes a student model on the resulting trajectories.
We introduce \papername{}, an autonomous post-training system
in which an LLM agent develops these interventions through
trajectory analysis, skill design, data selection, and feedback
from rollouts and evaluation.
Across Terminal, Search, and software engineering tasks with
4B and 9B student models, \papername{} improves mean performance
over Vanilla SFT on successful unguided teacher trajectories
by 1.42--8.99 percentage points, with the teacher, task pool,
training set size, and SFT recipe held fixed.
Together with the ablations and trajectory analyses, these results
suggest that behavioral shaping can make a fixed model
a more effective teacher.
We further find that retaining skill instructions during SFT
yields higher mean performance than removing them across all
six settings, despite their absence at evaluation and identical
supervised outputs.
These findings suggest that preserving skill context as part of
a training demonstration can support better behavioral internalization.
\end{abstract}

\makeatletter
\begin{document}

% --- hide footnote marker ONLY for title/affil footnote ---
\begingroup
\makeatletter
\renewcommand{\thefootnote}{}% no a/1/... marker
\renewcommand{\@makefnmark}{}% no mark in text
\maketitle
\makeatother
\endgroup

\begin{figure*}[h]
    \centering
    \includegraphics[width=\linewidth]{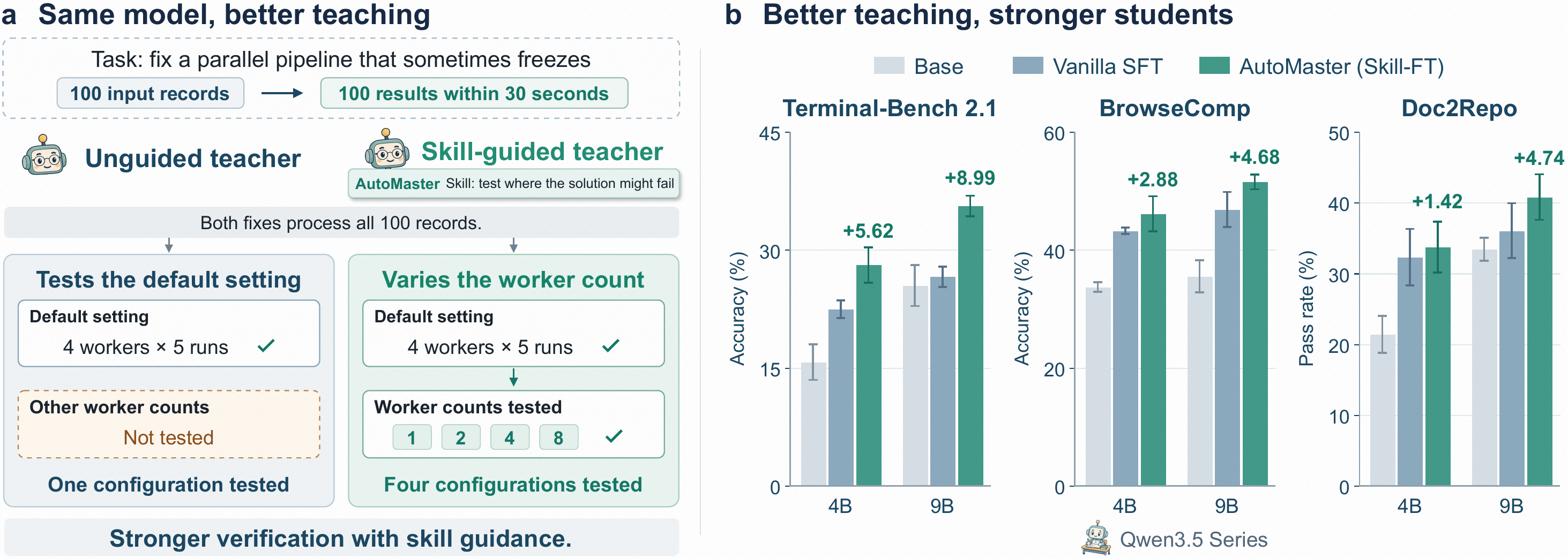}
    % \vspace{-1.5em}
    \caption{\textbf{A correct result is only part of what a teacher demonstrates.}
    \papername{} designs natural-language skills to guide teacher rollouts
    and selects demonstrations for fine-tuning.
    \textbf{(a)} In this example, both pipelines process all 100 records with four workers. The skill-guided teacher also checks whether the fix holds with fewer or more workers.
    \textbf{(b)} With the same teacher, task pool, training set size,
    and SFT recipe, students trained with \papername{} achieve higher
    performance than their Vanilla SFT counterparts on all three
    benchmarks at both model sizes (4B and 9B).
    Labels above the bars show gains of 1.42--8.99 percentage points
    over Vanilla SFT. Error bars show $\pm$1 SD over three evaluation
    runs per checkpoint.}
    \label{fig:teaser}
\end{figure*}

\section{Introduction}
LLM agents are beginning to undertake model development, from creating training data to running post-training pipelines~\citep{kulikov2026autodata,chen2026agent2,posttrainbench,zhao2026immersion}.
This opens a path toward autonomous post-training, in which an agent diagnoses model weaknesses, proposes interventions, and uses training feedback to decide what to try next.
A central question is what such a system should seek to change.

A common approach to post-training is supervised fine-tuning
(SFT) on successful teacher trajectories~\citep{chen2023fireact,ta2026proofgen}.
Yet \emph{task success does not specify which behaviors post-training
should reinforce}.
A teacher can reach the correct outcome despite unsupported assumptions, unproductive repetition, or superficial verification.
Fine-tuning on such successful trajectories can reinforce these
behaviors alongside the actions that support success.
When the teacher rarely demonstrates a useful behavior, even a large
pool of successful trajectories may offer few examples for the student
to learn from.
We therefore argue that autonomous post-training should actively shape how desirable demonstrations are generated as well as select which ones the student learns from.

To this end, we propose \textbf{\method{}} (Skill-Guided Fine-Tuning), which uses natural-language skills to shape teacher behavior during rollout.
A skill specifies how to reason or act in situations that recur across
tasks, such as checking a claim against independent evidence. 
We append the applicable skills to the system prompt to discourage undesirable habits and elicit useful behaviors that are rare in unguided rollouts.
We then fine-tune the student on selected teacher trajectories with
their original system prompts, including any skill instructions that
guided the teacher.

Developing these interventions involves two questions: whether a skill elicits the intended behavior from the teacher and how well the student learns from the resulting demonstrations. 
We introduce \textbf{\papername{}}, an autonomous post-training framework in which an LLM agent decides what the teacher should demonstrate and which trajectories the student should learn from.
The agent inspects trajectories to identify behaviors worth changing, designs skills for relevant tasks, and tests them in pilot teacher rollouts.
It uses these trials to refine the skills and their scope, then applies promising interventions to larger batches and selects demonstrations based on both task outcomes and observed behaviors.
In every round, SFT starts from the same initial student checkpoint. 
The agent uses rollout evidence and student evaluation to decide whether to revise a skill, change where it is applied, or select different demonstrations in the next round.

We evaluate \papername{} on terminal, search, and software engineering tasks with Qwen3.5-4B and Qwen3.5-9B~\citep{qwen2026qwen35} as students.
With the teacher, training task pool, number of demonstrations, and SFT recipe held fixed, \papername{} improves mean performance over Vanilla SFT on successful unguided teacher rollouts by 1.42--8.99 percentage points across the six settings.
Removing either skill guidance or agent-directed selection lowers performance in every setting.
The full method also outperforms SFT on unguided demonstrations matched by task ID and per-task sample count, showing that matching the training tasks alone does not recover the full gains.
These comparisons reveal \emph{post-training headroom that Vanilla SFT on successful unguided rollouts leaves untapped}. 
By using \method{} to shape what the teacher demonstrates and selecting which trajectories enter student training, \papername{} makes the same model a more effective teacher.

We further find that \emph{reducing a training--evaluation prompt mismatch can make the same trajectories less effective for learning}.
With the same trajectories and SFT recipe, retaining skill instructions during training yields higher performance than removing them across all six settings, even though all students are evaluated without generated skills.
The training examples suggest a possible explanation: retaining skill instructions keeps behavioral rules and their conditions of application explicit in the input.
Removing them preserves the same reasoning and actions as supervision, but may require the student to infer more of these rules from the task and trajectory.
For reasoning that explicitly references a skill, retaining the instructions also preserves the correspondence between the input guidance and the supervised output.
These findings suggest that preserving skill context can help the student better internalize the behaviors shaped during teacher rollouts.

Our contributions are summarized as follows:
\begin{itemize}
    \item \textbf{Behavioral shaping beyond task success.}
    We introduce \method{}, which uses natural-language skills to shape 
    the reasoning and actions a fixed teacher demonstrates during
    rollout, including on tasks it already solves.

    \item \textbf{Autonomous post-training framework.}
    We develop \papername{}, a framework in which an LLM agent designs, executes, and refines behavioral interventions.
    It jointly optimizes skill guidance and data selection using rollout evidence to guide subsequent exploration.

    \item \textbf{Skill context for behavioral internalization.}
    Controlled comparisons across the six settings show that retaining skill instructions in the SFT input improves student performance by 2.15--4.01 percentage points even without test-time skills. 
    This suggests that preserving the context behind supervised reasoning and actions can support better behavioral internalization.
\end{itemize}

\section{Related Work}
\label{sec:related_work}

\paragraph{Autonomous Post-Training.}
Recent systems use LLM agents to implement training experiments and revise them in light of the results~\citep{karpathy2026autoresearch,chen2026toward,wei2026autofinetune}.
FT-Agent~\citep{li2026ftdojo} uses structured planning and feedback analysis to refine data and training strategies, while RSI-Master~\citep{du2026rsimaster} uses persistent experiment records and reviewer feedback to guide exploration.
\papername{} focuses on two decisions: how to guide a fixed teacher
with skills and which of its demonstrations to use for student training.

\paragraph{Skill-Based Agent Learning.}
Prompt and skill optimization methods use execution feedback to refine natural-language guidance without updating model parameters~\citep{agrawal2025gepa,shi2026skillkd}.
Related work couples the evolution of skill libraries or procedural scaffolds with model training~\citep{xia2026skillrl,ding2026scaffold,ge2026skillgym}.
Skill-to-LoRA~\citep{zhang2026skilltolora} trains skill-specific adapters on skill-guided demonstrations, while Skill-SD~\citep{wang2026skillsd} uses a skill-conditioned teacher to
score tokens in the student's own rollouts for on-policy self-distillation.
% SkillGym~\citep{ge2026skillgym} transforms skill documents into executable, verifiable training environments.
Our work examines both how skills shape teacher demonstrations and whether retaining the skill instructions during SFT improves learning when those instructions are absent at evaluation.

\paragraph{Learning from Agent Trajectories.}
Prior work improves reasoning and agent capabilities by learning from generated trajectories~\citep{chen2024alphamath,chen-etal-2024-step,zeng2024agenttuning}, including trajectories that demonstrate error recovery~\citep{fu2025agentrefine,yuan2025agentr}.
Behavior Priming~\citep{jin2026beneficial} selects trajectories exhibiting beneficial behaviors, while PROOF-Gen~\citep{ta2026proofgen} uses per-task prompt optimization to recover verified demonstrations from teacher failures.
\method{} complements trajectory selection and failure recovery by actively shaping behavioral patterns during teacher rollouts, including on tasks it already solves.

\section{Autonomous Post-Training with \method{}}
\begin{figure*}[t]
    \centering
    \includegraphics[width=\linewidth]{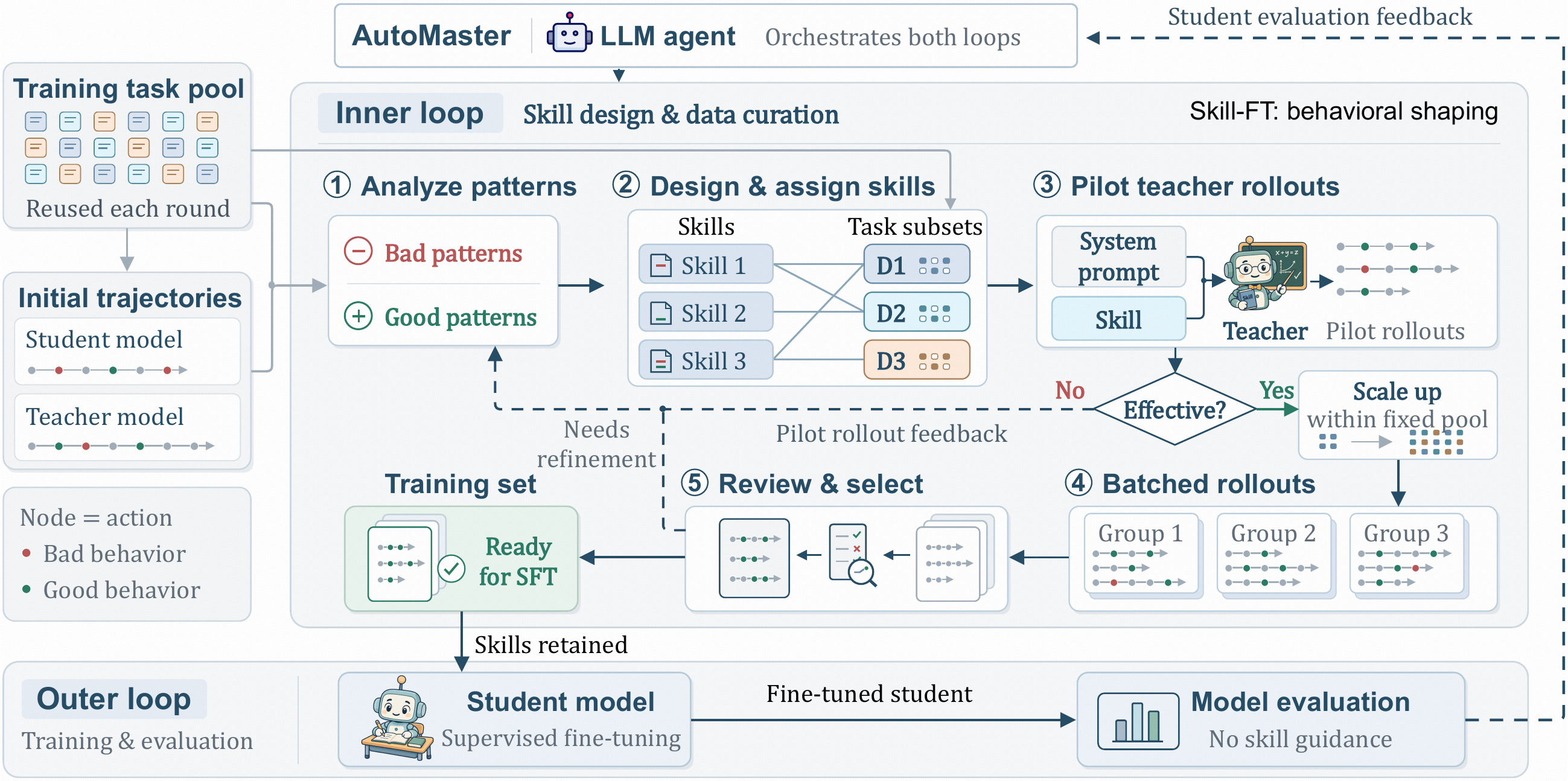}
    \caption{
    \textbf{Overview of \papername{}.}
    An LLM agent designs skills for subsets of a fixed training task pool and tests them in pilot teacher rollouts. It refines the skills as needed, runs promising interventions on larger batches, and selects demonstrations for SFT. In each outer-loop round, the student is trained from the same initial checkpoint, with any skill instructions used by the teacher retained in the training input. The student is evaluated without generated skills.
    }
    \label{fig:framework}
\end{figure*}

\papername{} uses an LLM agent to develop and refine \method{} interventions through two nested loops (Figure~\ref{fig:framework}). Within each setting, the teacher, training task pool, and initial student checkpoint are fixed. In the inner loop, the agent decides how to guide the teacher and which demonstrations to use for training. In the outer loop, it fine-tunes the student on the selected demonstrations and uses the student's subsequent evaluation scores and trajectories to guide the next round.

\subsection{\method{}: Skill-Guided Fine-Tuning}
\label{sec:skill_ft}
To shape the behavioral patterns represented in training trajectories, we guide the teacher with natural-language skills.
These skills are designed to suppress undesirable habits and elicit effective behaviors that are underrepresented in unguided rollouts.
A skill specifies how to reason or act in relevant situations and may include behavioral principles, concrete procedures, or domain-specific knowledge.
Skills can be shared across tasks or combined to address different aspects of the same task.
For a task $x$, let $s$ denote the combined text of the selected skills.
We append $s$ to the base system prompt $p$ and generate a trajectory through interaction with the task environment:
\begin{equation}
    \tau \sim \operatorname{Rollout}(\pi_T, x; p \oplus s),
    \label{eq:skill_rollout}
\end{equation}
where $\pi_T$ denotes the teacher model and $\oplus$ denotes text concatenation.
This guidance conditions the teacher's reasoning and actions during execution, enabling behavioral interventions even on tasks it already solves.

To internalize the demonstrated behaviors, we fine-tune the student on selected teacher trajectories using standard SFT.
For each trajectory, we retain the original system prompt, including any skill instructions, so the student sees the same instructions that guided the teacher.
At evaluation, the student follows the benchmark's official protocol without the generated skills.

\subsection{Inner Loop: Developing Behavioral Interventions}
\label{sec:inner_loop}
The agent begins by examining the training tasks and initial trajectories from both the unguided teacher and the initial student. It looks for habits to discourage and useful behaviors that the teacher rarely demonstrates, including on tasks the teacher already solves. In later rounds, the agent also inspects student evaluation trajectories to see where the intended behaviors appear and which weaknesses persist. These observations guide the skills it designs and the training tasks to which it applies them.

As shown in Figure~\ref{fig:framework}, pilot rollouts test whether candidate skills elicit the intended behaviors without introducing new problems during task execution.
The agent uses the resulting trajectories to refine both the skill instructions and their scope of application, repeating trials as needed.

The agent then scales promising interventions to batched rollouts and selects trajectories it judges valuable for training, considering both task outcomes and demonstrated behaviors.
The agent may also select unguided teacher trajectories that already exhibit useful behavioral patterns.
Feedback from generation and selection can prompt further revisions to skills or task coverage, followed by additional rollouts as needed.
The selected trajectories form the training set for the outer loop.

\subsection{Outer Loop: Model Training and Feedback}
\label{sec:outer_loop}
The outer loop evaluates the student trained on the demonstrations selected in each round.
In round $r$, the inner loop produces a training set $\mathcal{D}_r$.
SFT starts from the same initial student checkpoint $\theta_0$ in every round:
\begin{equation}
    \theta_r = \operatorname{SFT}(\theta_0, \mathcal{D}_r).
    \label{eq:outer_sft}
\end{equation}
The agent receives both scores and trajectories from evaluating $\theta_r$ without generated skills.
Scores summarize how well the student performs, while trajectories let the agent inspect how it handled individual tasks.
The agent can compare these observations with earlier skills, teacher rollouts, and training sets to decide whether to revise a skill, change its task coverage, or select different demonstrations in the next round. This experiment history carries forward, while each student is trained from $\theta_0$.

\begin{figure*}[t]
    \centering
    \includegraphics[width=0.95\linewidth]{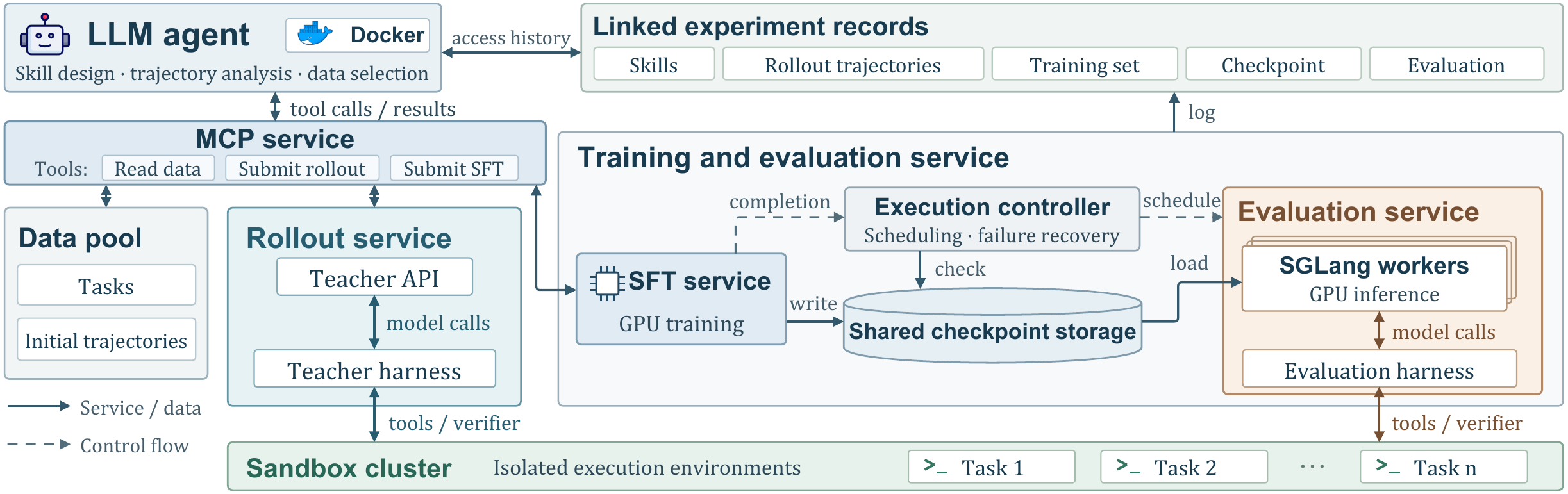}
    \caption{\textbf{Infrastructure of \papername{}.}
    The agent accesses data and submits rollout and SFT jobs through the MCP
    interface. After SFT, the execution controller automatically schedules
    evaluation on a separate GPU cluster, where SGLang workers load the
    checkpoint directly from storage shared with the training cluster.
    Linked experiment records let the agent trace each evaluation back to
    the training set and the skills that guided the teacher.}
    \label{fig:infra_framework}
\end{figure*}

\subsection{Infrastructure}
\label{sec:execution_infrastructure}

The infrastructure lets the \papername{} agent test how a skill shapes teacher
behavior and examine whether the student exhibits that behavior after
fine-tuning (Figure~\ref{fig:infra_framework}). 
From its Docker container, the agent uses the Model Context Protocol
(MCP)~\citep{anthropic2024mcp} to access tasks and initial trajectories,
submit teacher rollout jobs, and launch SFT on selected demonstrations.
The rollout harness calls the teacher API,
while the evaluation harness queries the student model served by
SGLang~\citep{zheng2024sglang}. Both harnesses use a shared sandbox cluster
that runs task tools and verifiers in isolated Docker containers, allowing
multiple tasks to execute concurrently.

When an SFT job completes, the execution controller checks that the
checkpoint is available and automatically schedules evaluation. The training
and evaluation GPU clusters share a checkpoint store: SGLang workers load
the weights directly from the location where the SFT service wrote them,
avoiding an intermediate copy to the evaluation cluster. The controller also
records job and stage progress so that it can restart a failed inference
worker and resume evaluation without rerunning SFT.

Across rounds, \papername{} links each student's evaluation scores and
trajectories to its checkpoint, its training set, and the skills used during
teacher rollouts. When a weakness recurs in student evaluations, the agent
can inspect earlier rollouts to see whether the teacher demonstrated the
desired behavior and check which of those demonstrations entered the
training set. This evidence informs whether the agent revises a skill,
changes its data selection, or both in the next round.
\section{Experiments}

\subsection{Experimental Setup}
\label{sec:experimental_setup}

\paragraph{Benchmarks and Task Pools.}
We evaluate Terminal, Search, and software engineering (SWE) tasks using Terminal-Bench 2.1~\citep{merrill2026terminalbench}, BrowseComp (BC; using the Lite-185 subset released by Traverse)~\citep{wei2025browsecomp,ma2026traverse}, and Doc2Repo~\citep{chen2026beyondswe}, respectively.
For each domain, we randomly sample 1,000 training tasks from public datasets~\citep{meng2026calibforge,deng2026fortsearcher,zhao2026denovoswe} to form a fixed task pool shared across all methods.

\paragraph{Models and Baselines.}
We use Qwen3.5-4B and Qwen3.5-9B~\citep{qwen2026qwen35} as students.
Across all experiments, the LLM agent runs in Codex with GPT-6 Astra~\citep{openai2026gpt6astra} at \texttt{xhigh} reasoning effort, while DeepSeek-V4-Pro-Preview~\citep{deepseek2026v4} serves as the teacher at \texttt{high} reasoning effort.
We compare against \textbf{Base}, the student without additional fine-tuning, and \textbf{Vanilla SFT}, which performs standard SFT on 512 trajectories randomly sampled from successful unguided teacher rollouts on the shared task pool.

\paragraph{Training and Evaluation.}
For \papername{}, the agent starts with trajectories generated
by the initial student and the unguided teacher on the
training task pool, with a budget of five outer-loop rounds.
Across all settings, each training run starts from the corresponding
initial checkpoint and performs full-parameter SFT on 512
trajectories for three epochs, with a learning rate of $10^{-5}$
and a global batch size of 32.
We evaluate final-epoch checkpoints without generated skills
using domain-specific harnesses~\citep{meng2026calibforge,sun2026aewm,chen2026beyondswe},
with identical settings across methods within each domain.
We measure task accuracy on Terminal-Bench 2.1 and BC and mean
test pass rate on Doc2Repo, reporting mean $\pm$ SD over three
evaluation runs per checkpoint.
Further experimental details are provided in
Appendix~\ref{app:experimental_details}.

\subsection{Main Results}
\label{sec:main_results}

% Requires: \usepackage{booktabs}
% Uses the manuscript's existing \papername macro.
\begin{table*}[t]
    \centering
    \caption{
        Main results with Qwen3.5-4B and Qwen3.5-9B.
        We report task accuracy for Terminal-Bench 2.1 and BrowseComp,
        and mean test pass rate for Doc2Repo, all in percent
        (higher is better).
        Results are mean $\pm$ SD over three evaluation runs per checkpoint.
        Bold marks the highest mean in each column.
        $\Delta$ gives the gain over Vanilla SFT in percentage points.
    }
    \label{tab:main_results}
    \small
    \setlength{\tabcolsep}{4pt}
    \renewcommand{\arraystretch}{1.16}
    \begin{tabular*}{\textwidth}{@{\extracolsep{\fill}}l*{6}{c}@{}}
        \toprule
        & \multicolumn{2}{c}{\textbf{Terminal-Bench 2.1}}
        & \multicolumn{2}{c}{\textbf{BrowseComp}}
        & \multicolumn{2}{c}{\textbf{Doc2Repo}} \\
        \cmidrule(lr){2-3}
        \cmidrule(lr){4-5}
        \cmidrule(lr){6-7}
        \textbf{Method}
        & \textbf{4B} & \textbf{9B}
        & \textbf{4B} & \textbf{9B}
        & \textbf{4B} & \textbf{9B} \\
        \midrule
        Base
        & $15.73 \pm 2.25$ & $25.47 \pm 2.59$
        & $33.69 \pm 0.82$ & $35.50 \pm 2.72$
        & $21.39 \pm 2.61$ & $33.42 \pm 1.60$ \\
        Vanilla SFT
        & $22.47 \pm 1.12$ & $26.59 \pm 1.30$
        & $43.24 \pm 0.54$ & $46.85 \pm 2.98$
        & $32.30 \pm 4.00$ & $36.05 \pm 3.89$ \\
        \addlinespace[2pt]
        \textbf{\papername{}}
        & $\mathbf{28.09} \pm 2.25$ & $\mathbf{35.58} \pm 1.30$
        & $\mathbf{46.13} \pm 2.98$ & $\mathbf{51.53} \pm 1.25$
        & $\mathbf{33.72} \pm 3.58$ & $\mathbf{40.79} \pm 3.21$ \\
        \midrule
        $\Delta$ over Vanilla SFT
        & $+5.62$ & $+8.99$
        & $+2.88$ & $+4.68$
        & $+1.42$ & $+4.74$ \\
        \bottomrule
    \end{tabular*}
\end{table*}

Table~\ref{tab:main_results} shows that \papername{} achieves the highest mean score across all six benchmark--model comparisons.
Vanilla SFT improves over Base in every setting, demonstrating the effectiveness of learning from successful teacher trajectories.
Building on this baseline, \papername{} yields further gains of 1.42--8.99 percentage points with the same training set size and SFT recipe.
Paired confidence intervals for these gains are reported in
Appendix~\ref{app:paired_performance}.
The gains are larger for the 9B model than for the 4B model across all three domains.

On Terminal-Bench 2.1, Vanilla SFT raises the 9B student's accuracy
from 25.47\% to 26.59\%. With the same teacher, training task pool,
number of demonstrations, and SFT recipe, \papername{} reaches 35.58\%.
\emph{The modest gain from imitating successful unguided rollouts
therefore understates what this student can learn from the same teacher}:
changing how demonstrations are generated and selected yields a further
gain of 8.99 percentage points in mean accuracy.

\begin{figure*}[t]
    \centering
    \includegraphics[width=\linewidth]{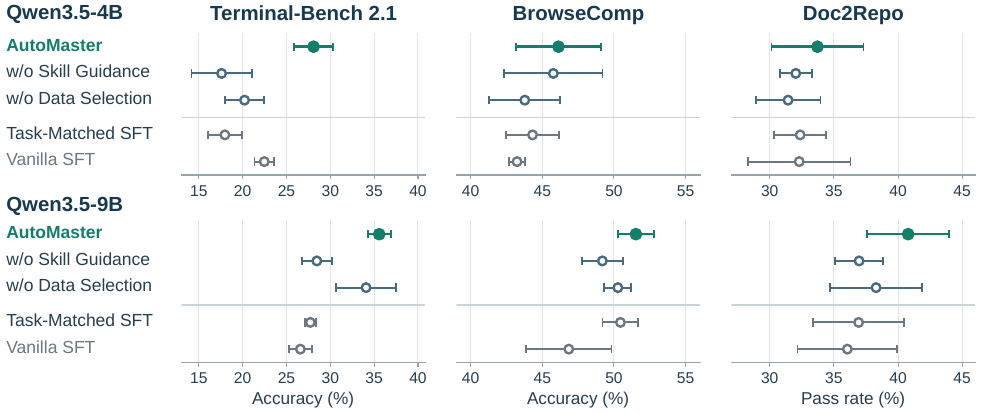}
    \vspace{-1.5em}
    \caption{\textbf{Contributions of skill guidance and data selection.}
    % Rows correspond to student model sizes and columns to benchmarks.
    \emph{w/o Skill Guidance} retains agent-directed selection from unguided
    teacher rollouts. \emph{w/o Data Selection} retains skill guidance on a
    fixed random subset of tasks but omits agent-directed trajectory selection.
    Removing either component lowers the mean score in each of the six settings.
    \emph{Task-Matched SFT} uses unguided rollouts from the same teacher and
    matches the final training set of \papername{}. Even with this match, \papername{} scores
    1.08--10.11 percentage points higher on average within each setting.
    Points show mean scores over three evaluation runs per checkpoint.
    Horizontal bars indicate $\pm$1 standard deviation.}
    % \caption{\textbf{Contributions of skill guidance and data selection.}
    % \emph{w/o Skill Guidance} retains agent-directed selection from unguided
    % teacher rollouts. \emph{w/o Data Selection} retains skill guidance on a
    % fixed random subset of tasks but omits agent-directed trajectory selection.
    % Removing either component lowers the mean score in all six settings.
    % \emph{Task-Matched SFT} uses unguided rollouts from the same teacher and
    % matches the task IDs and per-task sample counts of \papername{}'s final
    % training set. Points show means over three evaluation runs per checkpoint;
    % horizontal error bars indicate $\pm$1 SD.}
    \label{fig:component_ablations}
\end{figure*}

\subsection{Ablation Studies}
\label{sec:ablation_studies}

To understand how \papername{} improves on Vanilla SFT,
we separately ablate skill guidance and data selection.
In \textbf{w/o Skill Guidance}, the LLM agent selects training
trajectories from unguided teacher rollouts.
In \textbf{w/o Data Selection}, the agent designs and refines
skills for a fixed, randomly sampled subset of training tasks,
without agent-directed trajectory selection.
Both variants retain autonomous experimentation with the same
teacher, task pool, training set size, and SFT recipe.

Removing either component lowers mean performance in all six settings,
but their relative contributions vary
(Figure~\ref{fig:component_ablations}). On Terminal-Bench 2.1 with the
4B student, skill guidance without agent-directed selection yields
higher mean accuracy than selection from unguided rollouts
(20.22\% versus 17.60\%). Both fall below Vanilla SFT (22.47\%),
whereas the full method reaches 28.09\%. On BrowseComp at the same
model size, the ordering reverses: selection from unguided rollouts
yields 45.77\%, compared with 43.78\% for guidance without selection.
One possible explanation is that the value of selection depends on
what the teacher already demonstrates: selection can favor trajectories
that exhibit useful behaviors, while guidance can elicit behaviors
that are rare in unguided rollouts.

To test whether the choice of training tasks alone explains the gains,
we construct \textbf{Task-Matched SFT} from unguided rollouts by the
same teacher, matching both the task IDs and the per-task sample counts
of \papername{}'s final training set. With task composition held fixed,
\papername{} achieves higher mean scores in all six settings, by
1.08--10.11 percentage points. Across the four Terminal and Search
training corpora, both the unguided rollout and the selected
\papername{} demonstration succeed in 79.3--91.2\% of matched pairs
(Table~A5). For these pairs, the success label alone cannot distinguish
what the two trajectories demonstrate. Together with the component
ablations, these results support \emph{jointly optimizing how training
demonstrations are generated and selected}.

\section{Analysis}
\begin{figure*}[t]
    \centering
    \includegraphics[width=0.95\linewidth]{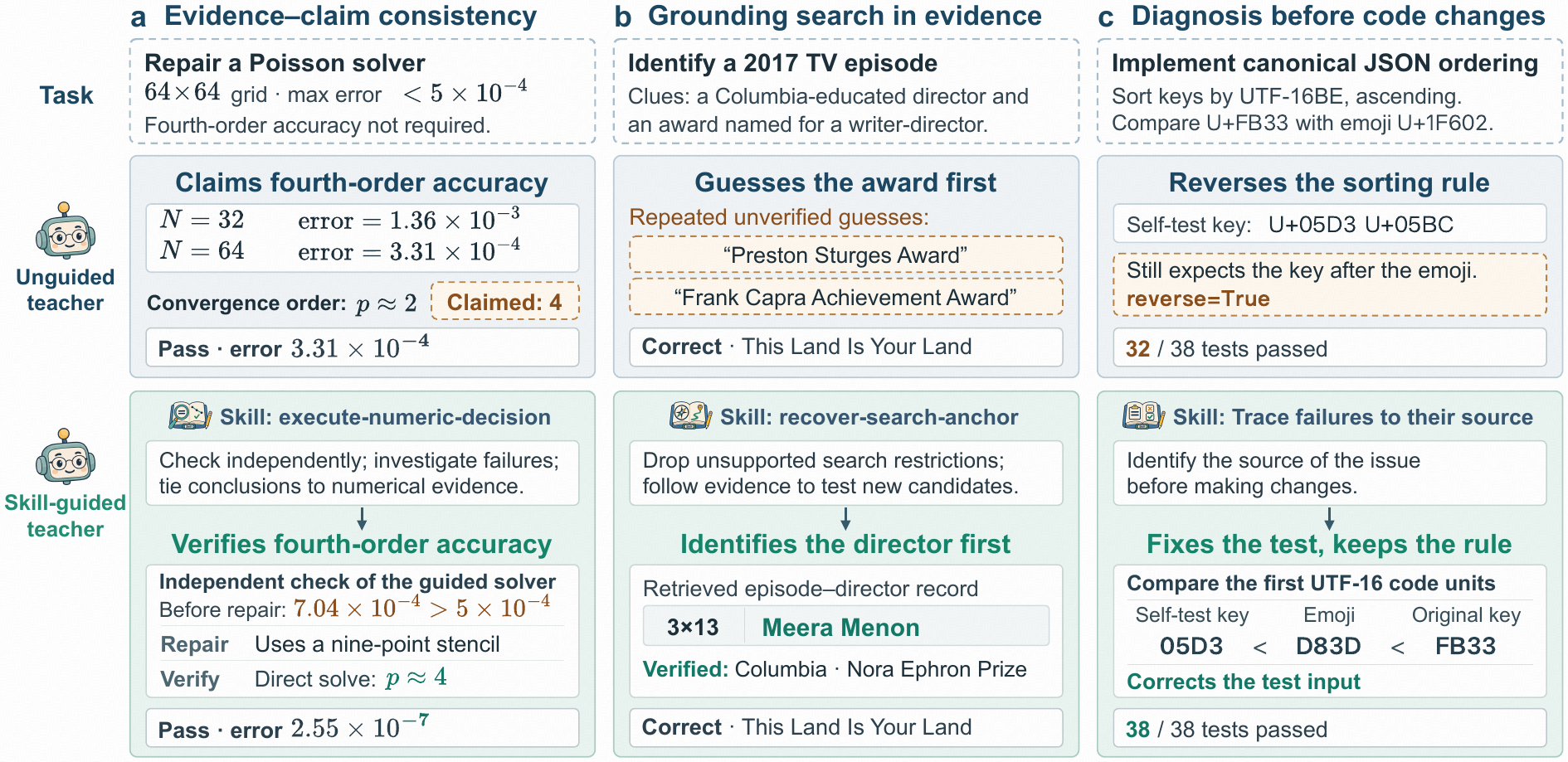}
    \caption{
        \textbf{Skill-guided changes in teacher behavior.}
        Each column compares unguided and skill-guided rollouts by the same
        teacher on the same task.
        \textbf{(a) Terminal:}
        Both solvers meet the error threshold. 
        The unguided teacher claims fourth-order accuracy, but an independent grid check of its solver indicates second-order convergence.
        The skill-guided teacher switches to a nine-point stencil and verifies fourth-order convergence using direct solves on successively finer grids.
        \textbf{(b) Search:}
        Both rollouts identify the same episode. The unguided teacher repeatedly
        guesses the award, while the skill-guided teacher identifies the director
        from episode credits and verifies the clues.
        \textbf{(c) SWE:}
        The unguided teacher reverses the global sort to satisfy a flawed self-test. The skill-guided teacher checks the UTF-16 encoding, corrects the self-test input, and keeps the required ascending order.
    }
    \label{fig:behavioral_shaping_cases}
\end{figure*}

\subsection{Behavioral Shaping Beyond Task Success}
\label{sec:behavioral_shaping}
Task success can conceal weaknesses in a teacher's reasoning and actions.
To examine the behaviors targeted by \method{}, Figure~\ref{fig:behavioral_shaping_cases} contrasts unguided and skill-guided teacher trajectories on the same tasks across three domains.

The first two cases illustrate behavioral differences that task outcomes alone cannot distinguish.
In Figure~\ref{fig:behavioral_shaping_cases}(a), both executions satisfy the numerical error threshold, yet the unguided teacher claims fourth-order accuracy despite measured convergence of approximately second order.
The guided teacher uses independent numerical checks to revise and validate its method.
Similarly, both search executions in Figure~\ref{fig:behavioral_shaping_cases}(b) reach the correct answer.
The unguided trajectory repeatedly pursues unverified award guesses before finding supporting records, while the guided trajectory uses retrieved episode--director records to identify and verify a candidate.
A filter based solely on successful completion would leave these behavioral differences unexamined.

The third case highlights the importance of diagnosing verification failures.
Both executions initially construct a self-test with an input different from the specification while retaining the original expected order.
The unguided teacher reverses the global sorting rule to satisfy this mistaken expectation.
The guided teacher checks the input encoding, corrects the self-test, and preserves the specified rule, passing 38 of 38 tests compared with 32 of 38.
The useful behavior here is to establish what a failed check actually implies before changing the implementation.

Across these cases, the guided trajectories illustrate changes in how
the teacher validates a numerical claim, redirects a search, or
diagnoses a failed check.
The first two cases show how skill guidance can address behavioral
weaknesses \emph{even when both executions succeed}, while the third
illustrates an improvement in implementation correctness.
These differences highlight the value of examining demonstrated
behaviors alongside task outcomes to identify opportunities
for improvement.
\papername{} identifies such opportunities through trajectory
analysis and designs skills to address them.
Together with the downstream results, these observations suggest
that \method{} can improve the training value of the demonstrations
produced by a fixed teacher.

\begin{figure*}[t]
    \centering
    \includegraphics[width=0.95\linewidth]{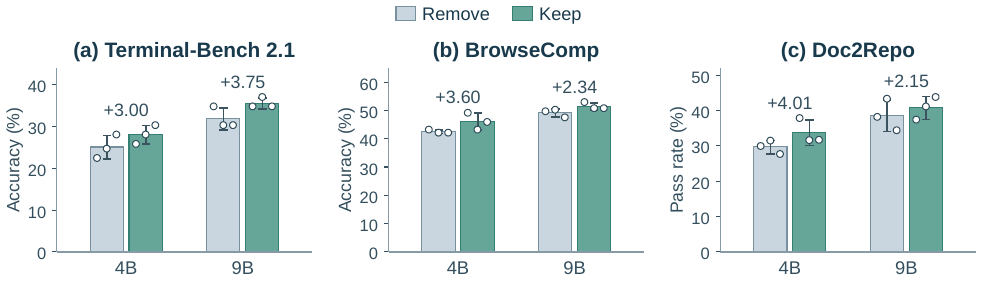}
    % \vspace{-1em}
    \caption{
        \textbf{Retaining skill instructions during SFT improves
        performance without test-time skills.}
        Both conditions use the same teacher trajectories and SFT recipe.
        Keep retains the skill instructions in the training input, whereas
        Remove omits them. All students are evaluated without skills under
        the official benchmark protocols.
        Bars show mean scores over three evaluation runs per checkpoint,
        circles show individual runs, and error bars indicate $\pm$1 SD.
        Labels give the gain of Keep over Remove in percentage points.
    }
    \label{fig:skill_internalization}
\end{figure*}
\begin{figure*}[t]
    \centering
    \includegraphics[width=\linewidth]{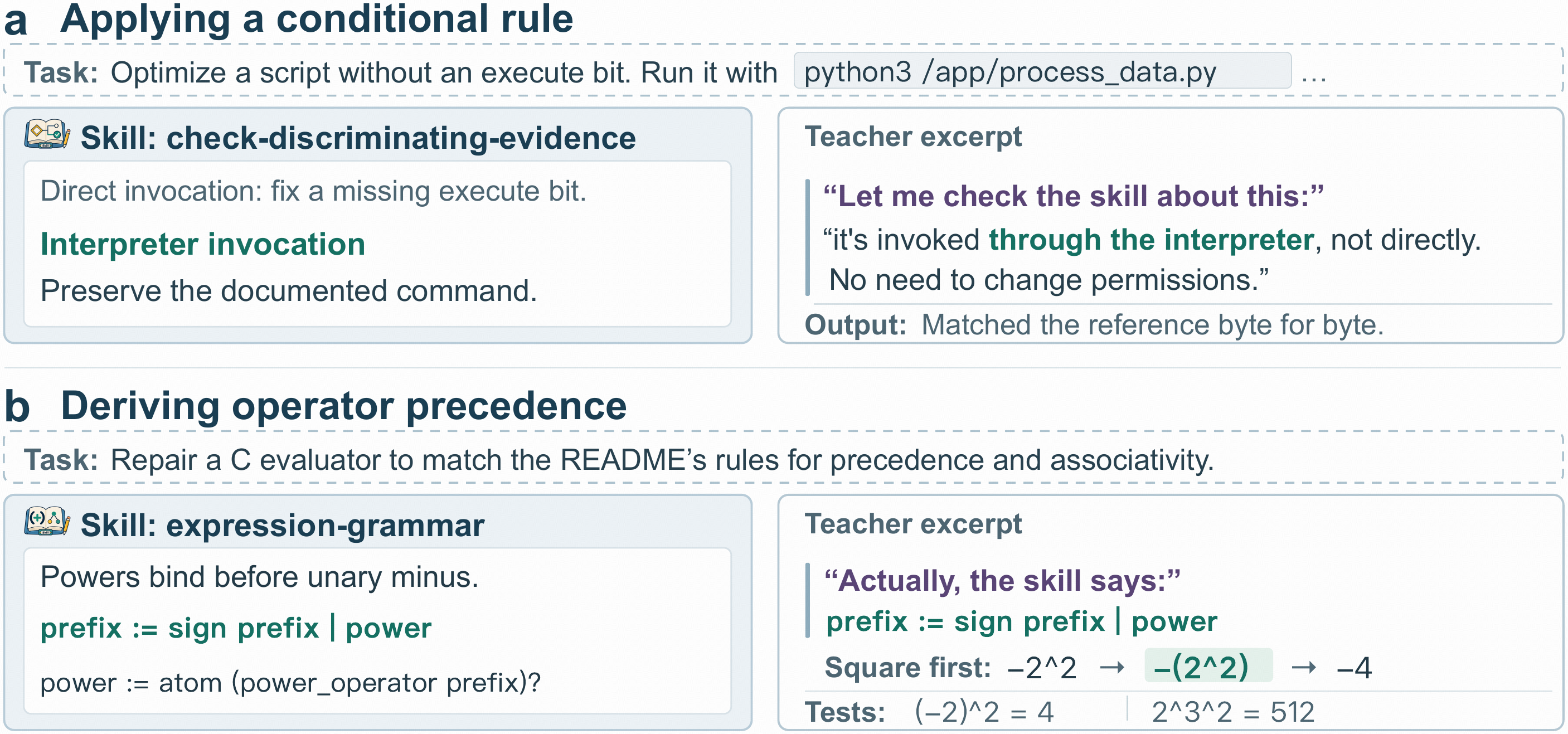}
    % \vspace{-1em}
    \caption{
        \textbf{Explicit use of skill context in teacher demonstrations.}
        Two examples from the SFT corpus illustrate how skill instructions
        inform the teacher's reasoning and actions.
        \textbf{(a)}
        The teacher applies a conditional rule to distinguish interpreter
        invocation from direct execution, avoiding an unnecessary permission change.
        \textbf{(b)}
        The teacher references the supplied grammar to derive an expression's
        grouping, then implements and tests the corresponding parser.
        Skill instructions and trajectory excerpts are condensed.
        Quotation marks indicate verbatim excerpts.
    }
    \label{fig:skill_context_training_cases}
\end{figure*}

\subsection{Preserving Skill Context for Better Internalization}
\label{sec:skill_internalization}

We investigate how to help the student internalize skill-guided behaviors.
Specifically, we examine whether the skill instructions used to guide
teacher rollouts should be retained during fine-tuning.
We compare \textbf{Keep}, which retains these instructions in the training
input, with \textbf{Remove}, which omits them.
Both conditions use the same trajectories and SFT recipe.
All models are evaluated without generated skills, following the official
evaluation protocols.

Removing the skill instructions leaves the supervised reasoning,
responses, and tool calls unchanged while making the training prompt
more like the evaluation prompt, which contains no generated skills.
Yet \emph{students perform better when the training input retains
the instructions that guided the teacher}. Across all six settings,
\textbf{Keep} exceeds \textbf{Remove} in mean score by 2.15--4.01
percentage points (Figure~6).

Figure~\ref{fig:skill_context_training_cases} illustrates a possible
explanation. In (a), the teacher consults the skill to decide whether
the script needs an execute bit: because the script runs through the
interpreter, no permission change is needed. In (b), the teacher explicitly
refers to the skill and uses its grammar to determine how to group an
expression. Removing the instructions preserves the teacher's reasoning
and actions as supervision, including any rule it quotes, but removes
the original instruction from the input. Retaining that context may
help the student learn when the demonstrated action applies.

\begin{figure*}[t]
    \centering
    \includegraphics[width=\linewidth]{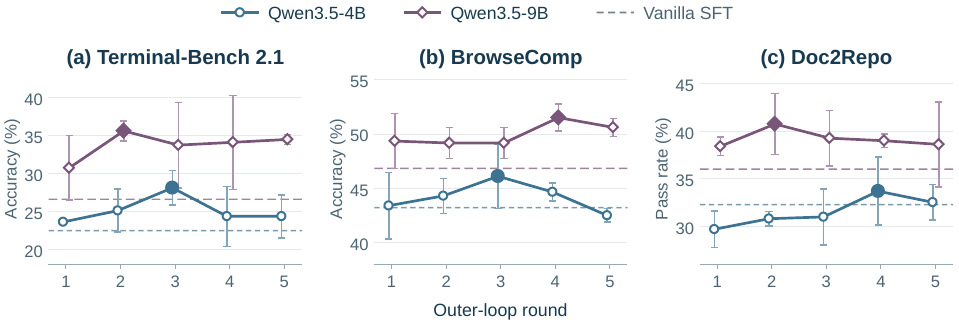}
    \caption{\textbf{Performance across five rounds of autonomous experimentation.}
    For each model size, every round trains a student from the
    initial checkpoint. Points show means over three evaluation runs per
    checkpoint; error bars indicate $\pm$1 SD. Dashed lines show the
    corresponding Vanilla SFT means. Filled markers indicate the highest
    observed mean for each model on each benchmark.}
    \label{fig:autonomous_experimentation}
\end{figure*}

\subsection{Autonomous Experimentation Across Rounds}
\label{sec:autonomous_experimentation}

Having examined how skill context affects learning from teacher
demonstrations, we next investigate how \papername{} develops
the interventions used to generate and curate those demonstrations.
The agent draws on accumulated rollout and training feedback
to explore alternative behavioral hypotheses and intervention
strategies.

Figure~\ref{fig:autonomous_experimentation} shows that continued
experimentation identifies higher-performing candidates in all
six benchmark--model settings.
The highest mean scores occur in rounds two through four,
exceeding first-round performance by 2.16--4.87 percentage points.
Variation across rounds is consistent with the exploratory protocol:
the agent continues to investigate alternative interventions after
finding a strong candidate, without requiring each trial to improve
on its predecessor.
On BrowseComp-9B, mean accuracy remains near 49\% for the first
three rounds before reaching 51.53\% in the fourth.
This delayed improvement illustrates how continued experimentation
can uncover further gains after several rounds of limited progress.

The consistency of gains also differs across model sizes.
The 9B model exceeds Vanilla SFT in mean performance in every round across all three domains, whereas the 4B model does so throughout all five rounds only on Terminal-Bench 2.1.
The component ablations show a similar pattern: both variants exceed Vanilla SFT in all three domains for the 9B model, but only on BrowseComp for the 4B model (Section~\ref{sec:ablation_studies}).
Together, these observations suggest that \emph{greater student capacity may support more consistent gains from the interventions explored by \papername{}}.

\section{Conclusion}

We introduced \method{} for shaping teacher behavior with natural-language skills and \papername{}, an autonomous post-training system that designs, executes, and refines these interventions.
Across three domains and two model sizes, \papername{} achieves higher mean performance than Vanilla SFT.
The results and trajectory analyses suggest that a fixed model can become a more effective teacher through behavioral shaping, even on tasks it already solves.
Retaining skill instructions during SFT also yields higher mean performance than removing them, despite their absence at evaluation, suggesting that preserving skill context can support better behavioral internalization.

\bibliography{main}
\clearpage
\appendix

\makeatletter
\@addtoreset{figure}{section}
\@addtoreset{table}{section}
\makeatother
\renewcommand{\thefigure}{\Alph{section}\arabic{figure}}
\renewcommand{\thetable}{\Alph{section}\arabic{table}}

\section{Experimental Details}
\label{app:experimental_details}

\subsection{Tasks and Evaluation Harnesses}
\label{app:tasks_and_harnesses}

Teacher rollouts and evaluation use an agent execution
framework built on AweAgent~\citep{aweagent2026}.
Table~\ref{tab:tasks_and_harnesses} summarizes the training sources,
evaluation benchmarks, and domain-specific harness references.
Each domain uses a fixed pool of 1,000 randomly sampled training
tasks shared across methods.
For BrowseComp, we evaluate on the Lite-185 subset released
by Traverse~\citep{ma2026traverse}.
For SWE, we use the harness of \citet{chen2026beyondswe}
with search tools disabled.

\begin{table*}[h]
    \centering
    \caption{
        Training sources, evaluation benchmarks, and harness references.
        Task counts refer to the evaluation sets.
    }
    \label{tab:tasks_and_harnesses}
    \small
    \setlength{\tabcolsep}{5pt}
    \renewcommand{\arraystretch}{1.15}
    \begin{tabularx}{\textwidth}{
        @{}l
        >{\raggedright\arraybackslash}X
        >{\raggedright\arraybackslash}X
        c
        >{\raggedright\arraybackslash}X@{}
    }
        \toprule
        \textbf{Domain}
        & \textbf{Training source}
        & \textbf{Evaluation benchmark}
        & \textbf{Tasks}
        & \textbf{Harness reference} \\
        \midrule
        Terminal
        & CalibForge~\citep{meng2026calibforge}
        & Terminal-Bench 2.1~\citep{merrill2026terminalbench}
        & 89
        & \citet{meng2026calibforge} \\

        Search
        & FORT-Searcher~\citep{deng2026fortsearcher}
        & BrowseComp-Lite~\citep{wei2025browsecomp,ma2026traverse}
        & 185
        & \citet{sun2026aewm} \\

        SWE
        & DeNovoSWE~\citep{zhao2026denovoswe}
        & Doc2Repo~\citep{chen2026beyondswe}
        & 50
        & \citet{chen2026beyondswe} \\
        \bottomrule
    \end{tabularx}
\end{table*}

\subsection{Teacher Rollout Configuration}
\label{app:teacher_rollout}

We use DeepSeek-V4-Pro-Preview~\citep{deepseek2026v4} for teacher rollouts across all domains, with the settings listed in Table~\ref{tab:teacher_rollout}.
Skill instructions, when applied, are appended to the base system prompt, and the teacher's reasoning content is retained in the interaction history.

\begin{table*}[h]
    \centering
    \caption{
        Teacher rollout settings across domains.
        Max completion tokens specifies the output budget
        for a single model response.
    }
    \label{tab:teacher_rollout}
    \small
    \setlength{\tabcolsep}{6pt}
    \renewcommand{\arraystretch}{1.15}
    \begin{tabularx}{\textwidth}{
        @{}>{\raggedright\arraybackslash}p{0.34\textwidth}
        *{3}{>{\centering\arraybackslash}X}@{}
    }
        \toprule
        \textbf{Parameter}
        & \textbf{Terminal}
        & \textbf{Search}
        & \textbf{SWE} \\
        \midrule
        Reasoning effort
        & \texttt{high}
        & \texttt{high}
        & \texttt{high} \\

        Temperature
        & 1.0
        & 1.0
        & 1.0 \\

        Top-$p$
        & 1.0
        & 1.0
        & 1.0 \\

        Max completion tokens
        & 16,384
        & 8,192
        & 16,384 \\

        Maximum execution steps
        & 500
        & 300
        & 400 \\

        Per-task time budget
        & 1 hour
        & No limit
        & 3 hours \\
        \bottomrule
    \end{tabularx}
\end{table*}

\subsection{Supervised Fine-Tuning}
\label{app:sft_configuration}

Table~\ref{tab:sft_configuration} summarizes the common SFT recipe used across methods, model sizes, and domains.
Each training run starts from the corresponding original student checkpoint, including successive outer-loop rounds of \papername{}.

We compute the autoregressive cross-entropy loss over assistant outputs, including reasoning, responses, and tool calls, and normalize it over supervised tokens.
System, user, and tool messages serve as context and are excluded from the loss.
By default, we retain each trajectory's original system prompt,
including any skill instructions used during generation.
In the \textbf{Remove} ablation
(Section~\ref{sec:skill_internalization}), we omit only the skill
instructions, leaving the remaining input context and supervised
outputs unchanged.

\begin{table}[t]
    \centering
    \caption{
        Shared SFT configuration across methods,
        student sizes, and domains.
    }
    \label{tab:sft_configuration}
    \small
    \setlength{\tabcolsep}{6pt}
    \renewcommand{\arraystretch}{1.12}
    \begin{tabularx}{0.85\linewidth}{
        @{}>{\raggedright\arraybackslash}X l@{}
    }
        \toprule
        \textbf{Parameter} & \textbf{Value} \\
        \midrule
        Training method & Full-parameter SFT \\
        Training trajectories & 512 \\
        Epochs & 3 \\
        Global batch size & 32 \\
        Maximum sequence length & 131,072 tokens \\
        \midrule
        Optimizer & Megatron Adam \\
        Learning rate & $10^{-5}$ \\
        Minimum learning rate & $10^{-6}$ \\
        Learning-rate schedule & Cosine decay \\
        Warmup ratio & 0.05 \\
        Adam $\beta_1$, $\beta_2$ & 0.9, 0.999 \\
        Weight decay & 0 \\
        Gradient clipping & 1.0 \\
        Training precision & \texttt{bfloat16} \\
        % Hardware & $8 \times$ H20 \\
        \bottomrule
    \end{tabularx}
\end{table}

\subsection{Model Evaluation}
\label{app:model_evaluation}

For each SFT run, we evaluate the checkpoint saved at the end
of the third epoch without generated skills.
We follow each benchmark's official evaluation protocol using
the domain-specific harnesses listed in
Table~\ref{tab:tasks_and_harnesses}.
Inference and execution settings are identical across methods
within each domain.
These benchmark evaluations provide scores and execution
trajectories for subsequent experimentation.

Across all domains, we use a temperature of 1.0, a top-$p$ of 0.95,
a maximum of 65,500 tokens per model response, and a limit of
600 execution steps per task.
The per-task agent time limit is three hours for Terminal-Bench 2.1
and four hours for Doc2Repo.
No overall task time limit is imposed for BrowseComp.

At the end of autonomous experimentation, \papername{} selects a
checkpoint based on the accumulated evaluation feedback and records
its choice in a final report. For all methods, we report the mean and
standard deviation over three evaluation runs per checkpoint. We use
task accuracy for Terminal-Bench 2.1 and BrowseComp and mean test pass
rate across tasks for Doc2Repo. The standard deviation reflects
evaluation variability for a fixed checkpoint.

\subsection{Detailed Ablation Results}
\label{app:ablation_results}

Table~\ref{tab:ablation_results} reports the complete numerical
results for the component ablations and Task-Matched SFT
comparison in Section~\ref{sec:ablation_studies}.

% Requires booktabs and the existing \papername macro.
\begin{table*}[t]
    \centering
    \caption{
        Component ablations and task-matched SFT results.
        Values are mean $\pm$ SD over three evaluation runs per
        checkpoint, in percent.
        We report accuracy for Terminal-Bench 2.1 and BrowseComp,
        and mean test pass rate for Doc2Repo.
        Bold marks the highest mean in each column.
    }
    \label{tab:ablation_results}
    \small
    \setlength{\tabcolsep}{4pt}
    \renewcommand{\arraystretch}{1.16}
    \begin{tabular*}{\textwidth}{@{\extracolsep{\fill}}l*{6}{c}@{}}
        \toprule
        & \multicolumn{2}{c}{\textbf{Terminal-Bench 2.1}}
        & \multicolumn{2}{c}{\textbf{BrowseComp}}
        & \multicolumn{2}{c}{\textbf{Doc2Repo}} \\
        \cmidrule(lr){2-3}
        \cmidrule(lr){4-5}
        \cmidrule(lr){6-7}
        \textbf{Method}
        & \textbf{4B} & \textbf{9B}
        & \textbf{4B} & \textbf{9B}
        & \textbf{4B} & \textbf{9B} \\
        \midrule
        \textbf{\papername{}}
        & $\mathbf{28.09} \pm 2.25$
        & $\mathbf{35.58} \pm 1.30$
        & $\mathbf{46.13} \pm 2.98$
        & $\mathbf{51.53} \pm 1.25$
        & $\mathbf{33.72} \pm 3.58$
        & $\mathbf{40.79} \pm 3.21$ \\
        w/o Skill Guidance
        & $17.60 \pm 3.43$
        & $28.46 \pm 1.72$
        & $45.77 \pm 3.43$
        & $49.19 \pm 1.43$
        & $32.03 \pm 1.24$
        & $36.97 \pm 1.88$ \\
        w/o Data Selection
        & $20.22 \pm 2.25$
        & $34.08 \pm 3.43$
        & $43.78 \pm 2.48$
        & $50.27 \pm 0.94$
        & $31.43 \pm 2.53$
        & $38.29 \pm 3.58$ \\
        \midrule
        Task-Matched SFT
        & $17.98 \pm 1.95$
        & $27.72 \pm 0.65$
        & $44.32 \pm 1.87$
        & $50.45 \pm 1.25$
        & $32.37 \pm 2.01$
        & $36.93 \pm 3.53$ \\
        Vanilla SFT
        & $22.47 \pm 1.12$
        & $26.59 \pm 1.30$
        & $43.24 \pm 0.54$
        & $46.85 \pm 2.98$
        & $32.30 \pm 4.00$
        & $36.05 \pm 3.89$ \\
        \bottomrule
    \end{tabular*}
\end{table*}

We construct Task-Matched SFT from the initial unguided
teacher rollouts corresponding to \papername{}'s final training
tasks.
Task IDs and per-task sample counts are preserved.
We retain both successful and unsuccessful unguided trajectories
to preserve the matched task composition.
Each trajectory retains its original unguided prompt,
without generated skill instructions.

Table~\ref{tab:paired_training_outcomes} characterizes the
paired teacher demonstrations in these training corpora.
Each row summarizes 512 pairs of demonstrations for the corresponding student.
For Terminal and Search, we compare binary task outcomes.
Across these four corpora, 8.79--20.70\% of the pairs contain
a successful \papername{} demonstration and an unsuccessful
unguided rollout.
The remaining pairs succeed in both conditions.
For SWE, we report whether the task score increases, remains
unchanged, or decreases under \papername{}, allowing partial
improvements and regressions to be represented.
These statistics describe the final selected training corpora
and complement the downstream student results in
Table~\ref{tab:ablation_results}.

% Requires booktabs and the existing \papername macro.
\begin{table*}[t]
    \centering
    \caption{
        Paired outcomes of teacher training demonstrations.
        $N$ denotes the number of paired training examples.
        The Student column gives the size of the student trained on each corpus.
        \textbf{(a)} Transitions are from the unguided teacher
        rollout to the \papername{} demonstration.
        \textbf{(b)} SWE categories compare task scores:
        improved, unchanged, and decreased indicate higher,
        equal, and lower scores under \papername{}, respectively.
    }
    \label{tab:paired_training_outcomes}
    \small
    \setlength{\tabcolsep}{4pt}
    \renewcommand{\arraystretch}{1.16}

    \begin{tabular*}{\textwidth}{@{\extracolsep{\fill}}llrrrrr@{}}
        \toprule
        \multicolumn{7}{@{}l}{
            \textbf{(a) Binary task outcomes in Terminal and Search}
        } \\
        \addlinespace[3pt]
        \textbf{Domain}
        & \textbf{Student}
        & $N$
        & \textbf{Failure $\rightarrow$ success}
        & \textbf{Both succeed}
        & \textbf{Success $\rightarrow$ failure}
        & \textbf{Both fail} \\
        \midrule
        Terminal & 4B & 512 & 106 & 406 & 0 & 0 \\
        Terminal & 9B & 512 & 100 & 412 & 0 & 0 \\
        Search   & 4B & 512 & 45  & 467 & 0 & 0 \\
        Search   & 9B & 512 & 64  & 448 & 0 & 0 \\
        \bottomrule
    \end{tabular*}

    \vspace{7pt}

    \begin{tabular*}{\textwidth}{@{\extracolsep{\fill}}lrrrr@{}}
        \toprule
        \multicolumn{5}{@{}l}{
            \textbf{(b) Task-score changes in SWE}
        } \\
        \addlinespace[3pt]
        \textbf{Student}
        & $N$
        & \textbf{Improved}
        & \textbf{Unchanged}
        & \textbf{Decreased} \\
        \midrule
        4B & 512 & 43 & 433 & 36 \\
        9B & 512 & 24 & 483 & 5  \\
        \bottomrule
    \end{tabular*}
\end{table*}

\subsection{Paired Analysis of Performance Gains}
\label{app:paired_performance}

We quantify uncertainty in the gains over Vanilla SFT using
paired bootstrap resampling~\citep{koehn2004statistical,dror2018hitchhikers}.
This estimates uncertainty in the performance difference while
preserving the correspondence between the two methods' scores
on each evaluation task.
For each task, we first average its scores over the three
evaluation runs for each method.
We then resample task IDs with replacement, retaining both
methods' scores for each sampled task, and recompute the mean
performance difference.
We use 10,000 resamples with random seed 0 to obtain pointwise
95\% percentile bootstrap confidence intervals.
For Doc2Repo, resampling is performed over whole repositories,
each corresponding to one task in the 50-repository evaluation set.
All tasks and their recorded scores are retained.
The intervals summarize variation across tasks or repositories
for the reported checkpoints and recorded evaluations.

\begin{table}[h]
    \centering
    \caption{
        \textbf{Paired performance gains over Vanilla SFT.}
        $\Delta$ denotes the mean gain of \papername{} over Vanilla SFT.
        Gains and pointwise 95\% bootstrap confidence intervals
        are reported in percentage points and computed from
        unrounded scores.
        $N$ denotes the number of evaluation tasks, or
        repositories for Doc2Repo.
    }
    \label{tab:paired_performance}
    \small
    \setlength{\tabcolsep}{4pt}
    \renewcommand{\arraystretch}{1.12}
    \begin{tabular}{@{}llrrc@{}}
        \toprule
        \textbf{Benchmark}
        & \textbf{Model}
        & $\boldsymbol{N}$
        & $\boldsymbol{\Delta}$
        & \textbf{95\% CI} \\
        \midrule
        Terminal-Bench 2.1
        & 4B & 89 & $+5.62$ & $[1.12,\ 10.49]$ \\
        & 9B & 89 & $+8.99$ & $[2.25,\ 16.10]$ \\
        \addlinespace[2pt]
        BrowseComp
        & 4B & 185 & $+2.88$ & $[-1.08,\ 6.85]$ \\
        & 9B & 185 & $+4.68$ & $[0.72,\ 8.83]$ \\
        \addlinespace[2pt]
        Doc2Repo
        & 4B & 50 & $+1.42$ & $[-1.94,\ 5.33]$ \\
        & 9B & 50 & $+4.74$ & $[0.65,\ 8.93]$ \\
        \bottomrule
    \end{tabular}
\end{table}

Table~\ref{tab:paired_performance} shows positive mean gains
across all six benchmark--model settings.
For the 4B model, the confidence interval lies entirely above
zero on Terminal-Bench 2.1, while the smaller mean gains on
BrowseComp and Doc2Repo have intervals that include zero.
For the 9B model, mean gains range from 4.68 to 8.99 percentage
points, and the lower bounds of all three 95\% confidence
intervals are positive, providing consistent paired evidence
of improvements across domains.

\section{Skills Generated by \papername{}}
\label{app:generated_skills}

This section presents four skills generated by \papername{} and used
in the teacher trajectories analyzed in
Sections~\ref{sec:behavioral_shaping} and~\ref{sec:skill_internalization}.
They illustrate how behavioral objectives are translated into
instructions specifying conditions of application, concrete
procedures, and domain-specific knowledge.

{\small\color{AMSkillMuted}
Skill instructions are reproduced verbatim, with formatting adapted
and metadata and enclosing tags omitted.
The numerical and grammar instructions are shown in full.
\texttt{[...]} marks omitted passages in the search and contract
instructions.\par}

\noindent\textbf{Evidence--claim consistency.}
This skill links completion claims to executable numerical checks
and requires the specified acceptance criterion to be preserved
during repair
(Figure~\ref{fig:behavioral_shaping_cases}(a)).

\begin{AMskillbox}[label={skill:execute-numeric-decision}]{execute-numeric-decision}{Complete instruction}
\textbf{Execute the numerical decision}

When correctness depends on numerical accuracy or convergence, put the relevant acceptance condition into the check you run. Adapt this finite-scalar example to the task's actual error convention:

\begin{Verbatim}[fontsize=\small,baselinestretch=1.0]
error = abs(actual - reference)
passed = math.isfinite(error) and error <= allowed_error
print(actual, reference, error, allowed_error, "PASS" if passed else "FAIL")
assert passed, "numerical requirement remains unresolved"
\end{Verbatim}

Define the reference and bound before running this block. Use the specified absolute, relative, or norm error and strictness; the example does not authorize substituting absolute error for a relative requirement. Keep prescribed input scale and iteration budget. If no bound is specified, justify a diagnostic bound and label it as such. An existing test that already makes the decision suffices. Returning a comparison from a test function does not enforce it: its result must reach the assertion or failure branch.

Base the reference on an independent analytic case, trusted implementation, or mathematical identity. Preserve exact comparisons for discrete results. Treat zero references and expected NaN/Inf according to the interface; apply finite-value checks only where finite output is required.

If the check fails, retain that case and condition while investigating. Save a random failing input or its reproducible seed. A common scale error suggests units, normalization or weights; an error near the bound suggests stopping criteria or error estimates. Inspect the responsible computation, make a justified correction, and rerun the same condition. Correct a mistaken reference only from independent evidence. Do not widen a required tolerance or replace a failing input with a passing one.

Connect completion to the observed decision, or state the unresolved requirement. More suites and repeated unchanged runs are unnecessary.
\end{AMskillbox}

\noindent\textbf{Evidence-guided search.}
When an identification search stalls, this skill ties the next
search action to the unsupported assumption or unresolved relation
(Figure~\ref{fig:behavioral_shaping_cases}(b)).

\begin{AMskillbox}[label={skill:recover-search-anchor}]{recover-search-anchor}{Instruction excerpts}
\textbf{Recover a stalled search anchor}

Use this when searches keep returning the same weak matches, or a promising candidate fits one clue but conflicts with another. The aim is to obtain evidence that can change the candidate, not to accumulate more mentions of it. If a search is yielding new relevant evidence, continue it.

\AMskillOmission
Choose the next action from the observed problem:

\noindent\textbullet\enspace \textbf{The candidate or scope is still a guess.} Drop that guessed restriction from the next discovery query. Search a different distinctive relation from the question that can identify an entity independently: an unusual office and succession, an event and venue, an original name and later renaming, or a study's measurable design. Use the clue's concrete terms, not every clause of the puzzle. If a new candidate appears, test the relation that the old candidate could not explain.

\AMskillOmission
At a consequential change, briefly connect the observation to the next action: what the results failed to establish, which assumption you are relaxing, and what the next query can distinguish. The actual query should implement that change; saying “try another approach” while retaining the same unsupported restriction is not a pivot. \AMskillOmission
\end{AMskillbox}

\noindent\textbf{Diagnosing verification failures.}
This skill pairs a check derived from the specification with
a contrasting input that distinguishes plausible implementations.
It requires diagnosing failures before changing the implementation
or the expected result. Figure~\ref{fig:behavioral_shaping_cases}(c)
shows \texttt{contract-checkpoint} under the descriptive title
\emph{Trace failures to their source}.

\begin{AMskillbox}[label={skill:contract-checkpoint}]{contract-checkpoint}{Instruction excerpts}
\textbf{Keep a runnable public contract through construction}

Apply this when parsing, conversion, serialization, configuration, or stateful APIs compose several operations into a visible result. For a small independent function, use a direct input/output check; do not invent an architecture or a large test suite.

\AMskillOmission
\textbf{Diagnose and preserve the failing example}

Read the actual failing assertion and its input before changing code. Reduce that case and inspect the type and value at the first boundary that could explain the mismatch. Check whether the error belongs to parsing, conversion, state, or output only as the observations require. Fix the shared cause and rerun the original public assertion plus its contrasting case.

If your expected value is wrong, establish why from the specification or a small independent calculation before changing it. When an example contradicts a general rule, record the conflict, test a second discriminating input, and choose a consistent interpretation; do not invent a new global algorithm merely to fit that example or claim the contradiction is resolved by a passing self-test.

Keep the same runnable script across repairs. Invoke it again after changes to shared helpers, imports, constructors, or state transitions. Add a small regression for a newly found mechanism instead of repeatedly rewriting a broad inline suite. Run it after the last relevant edit, alongside the task's required verification, and distinguish exercised behavior from paths that remain unverified.
\end{AMskillbox}

\noindent\textbf{Explicit rules and their scope.}
This skill supplies a grammar together with compatibility constraints
and discriminating tests.
The teacher explicitly consults this grammar when deriving expression
grouping (Figure~\ref{fig:skill_context_training_cases}(b)).

\begin{AMskillbox}[label={skill:expression-grammar}]{expression-grammar}{Complete instruction}
When implementing arithmetic with unary signs and powers, derive the grouping rules from the specification before changing parser calls. For conventional mathematical binding, recursive descent can use:

\begin{Verbatim}[fontsize=\small,baselinestretch=1.0]
sum     := product (("+" | "-") product)*
product := prefix (("*" | "/" | "%") prefix)*
prefix  := sign prefix | power
power   := atom (power_operator prefix)?
\end{Verbatim}

Here the power base excludes an ungrouped sign, but its right operand admits signs and further powers. In a Pratt or precedence-climbing parser, keep prefix parsing available at every operand entry; use infix binding thresholds for associativity. Adapt the existing parser, including only supported operators. For binary compatibility, preserve the reference's grouping instead of imposing mathematical conventions.

Check the actual evaluator with a signed base versus a grouped signed base, a signed expression in the right operand, and repeated powers. Derive expected values first; include a positive exponent formed with repeated signs when integer powers are required. Build-only repairs need no grammar change.
\end{AMskillbox}

These instructions guide teacher execution and, when retained
during SFT, form part of the input accompanying the supervised
reasoning and actions.
Their role as training context is examined in
Section~\ref{sec:skill_internalization}.

\end{document}